\documentclass[runningheads]{llncs}

\usepackage{eccv}

\usepackage{eccvabbrv}

\usepackage{algorithm}
\usepackage{algpseudocode}
\usepackage{mathtools}

\newcommand{\modelname}{Perceptio\xspace}

\usepackage{graphicx}
\usepackage{booktabs}
\usepackage{multirow}
\usepackage{lmodern}
\usepackage{wrapfig}

\usepackage{hyperref}

\usepackage{booktabs}
\usepackage{tabularx}
\newcommand{\orcidlink}[1]{}

\begin{document}

\title{\modelname: Perception Enhanced Vision Language Models via Spatial Token Generation}

\titlerunning{Perceptio: Perception Enhanced VLMs via Spatial Token Generation}

\author{Yuchen Li\inst{1} \and Amanmeet Garg\inst{1} \and
Shalini Chaudhuri\inst{1} \and
Rui Zhao\inst{1} \and Garin Kessler\inst{1}
}

\authorrunning{Y.~Li et al.}

\institute{Amazon\\
\email{}}
\maketitle
\begin{abstract}
Large Vision–Language Models (LVLMs) excel at semantic understanding but struggle with fine-grained spatial grounding, as the model must implicitly infer complex geometry without ever producing a spatial interpretation. We present \textit{Perceptio}, a perception-enhanced LVLM with 2D–3D spatial reasoning abilities, enabled via explicit semantic-segmentation tokens and depth tokens generated directly within the autoregressive sequence. Concretely, we (i) distill a VQ-VAE depth codebook from a strong monocular teacher to tokenize dense depth into compact sequences, and (ii) integrate SAM2-based semantic-segmentation tokens and VQ-VAE depth tokens inside the LLM so the model first emits spatial tokens and then answers. To stabilize depth token generation, we introduce novel composite depth-token objectives (marker, token, and count losses) and a soft-merging technique for differentiable reconstruction. We adopt a multi-task co-training strategy across diverse datasets, letting the model learn perception tokens to tackle multiple downstream tasks. Building on InternVL, Perceptio achieves state-of-the-art performance across benchmarks: improving referring expression segmentation by +0.8/+1.4/+1.1 cIoU on RefCOCO/+/g, HardBLINK spatial understanding accuracy by 10.3\%, and MMBench accuracy by 1.0\%, demonstrating that explicit spatial chain-of-thought materially strengthens spatial grounding in LVLMs.

\end{abstract}

\section{Introduction}
\label{sec:intro}


Modern open-source LVLMs such as the InternVL series \cite{chen2024expanding} and the Qwen-VL series \cite{bai2023qwenvl,wang2024qwen2vl} have scaled up vision backbones and introduced advanced alignment pipelines. These often deliver strong performance on tasks requiring multi-modal understanding such as captioning \cite{Xu2015ShowAA}, visual question answering (VQA) \cite{agrawal2016vqavisualquestionanswering}, and grounding \cite{Xiao2017WeaklySupervisedVG}.
Despite pre-training with web-scale image-text data, LVLMs often struggle with spatial understanding in images, including reasoning about depth, distance, and scale \cite{fu2024blinkmultimodallargelanguage,tong2024eyes}. For example, BLINK \cite{fu2024blinkmultimodallargelanguage} evaluated popular LVLMs on simple tasks that humans solve ``within a blink'' and observed that LVLMs barely surpass random guessing. This phenomenon is partly due to the lack of explicit 3D cues during pre-training, which also suggests robust \emph{spatial intelligence}—the ability to comprehend relative positions and spatial arrangements—has not yet emerged as a general skill. 
These findings motivate a design that can incorporate spatial understanding into the model learning.


\begin{wrapfigure}{r}{0.65\linewidth} 
    \centering
    \vspace{-1pt} 
    \includegraphics[width=\linewidth]{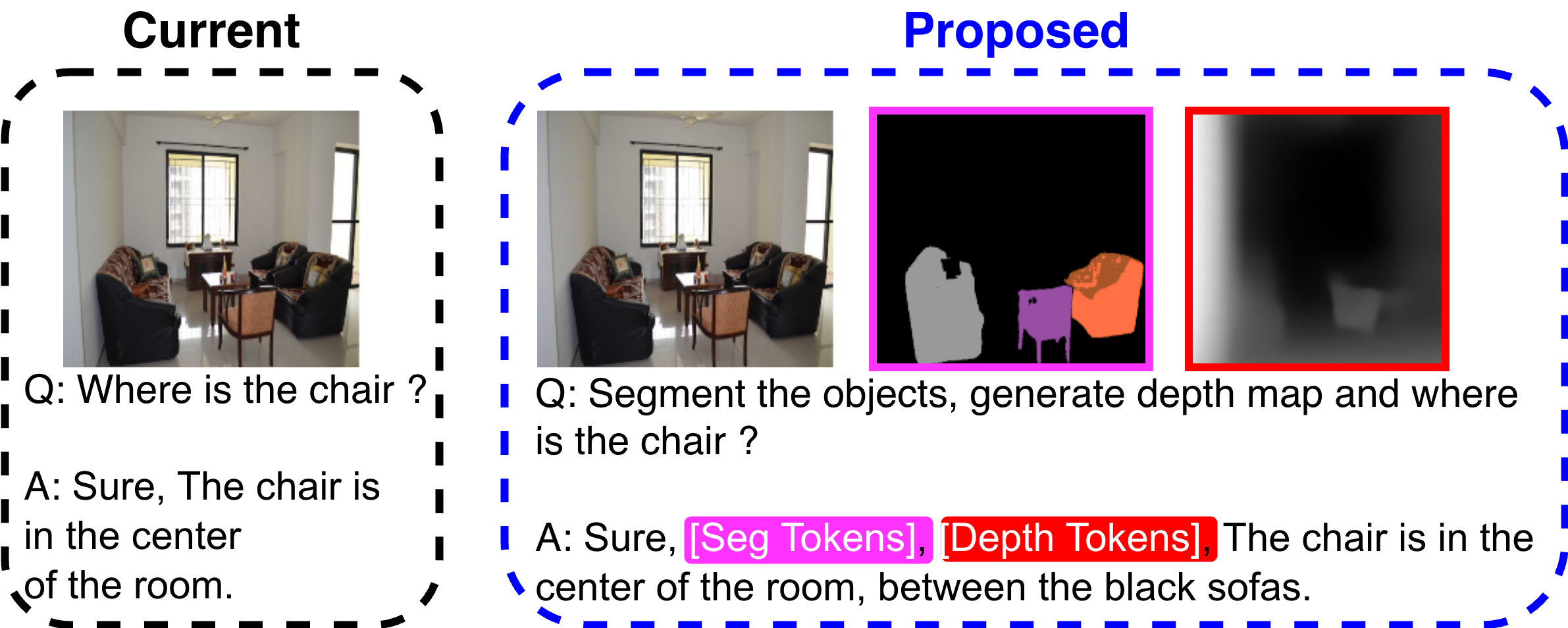}
    \caption{Comparison of Perceptio pipeline vs standard VLMs.}
    \label{fig:compared}
    \vspace{-20pt} 
\end{wrapfigure}

To address spatial understanding challenge for LVLM, we propose \textbf{\modelname}, a perception enhanced LVLM that \emph{jointly learns} to generate tokens for 2D semantic segmentation and 3D depth perception as an auto-regressive sequence. Building on InternVL-2.5 \cite{chen2024expanding}, Segment Anything Model 2 (SAM2) \cite{ravi2024sam2}, and Depth Anything V2 model \cite{yang2024depth}, \textbf{\modelname} emits a dedicated segmentation token and a depth token stream before producing the text token. This design enables a perception-enhanced conditional generation, where, by generating segmentation and depth tokens first, the model anchors subsequent language in explicit 2D \& 3D cues, improving VQA, grounding, and spatial reasoning.

We endow the LVLM with 3D spatial perception knowledge by distilling from a 3D depth generation model as teacher in a teacher–student framework. We train a Vector Quantized-Variational Autoencoder (VQ-VAE) on depth maps predicted by the specialist {Depth Anything V2} model \cite{yang2024depth}. Such a discretized depth token sequence and the resulting codebook indices serve as 3D perception tokens. We impart 2D spatial knowledge by incorporating a learnable segmentation token conditioned on the query text. We treat segmentation and depth as priors that condition the language decoder. In the standard setup, a text-only query $q$ maps to an answer $a$. In our setting, we augment the input with structured priors over the query and answer, as show in Figure~\ref{fig:compared}, formatted as 
\vspace{-5pt}
\begin{align*}
    [\texttt{seg tokens}],[\texttt{depth tokens}],[\texttt{text tokens}]
\end{align*}
With this perception-enhanced design, the model first interprets the perceptual signal, enabling more effective answers on the downstream task. 
We highlight our contributions by four main points:
\begin{enumerate}
    \item \textbf{Explicit spatial perception in LVLMs.} We introduce \modelname, which enhances LVLMs with in-sequence 2D segmentation and discretized 3D depth tokens, enabling pixel-level and geometric reasoning. To the best of our knowledge, \modelname is the first to jointly optimize for 2D and 3D perception signals within a single autoregressive sequence in LVLMs.
    \item \textbf{Unified Multi-task Training with Novel Depth Objectives.} We propose a joint text–segmentation–depth objective and a series of novel depth-token loss functions (marker + token + count) that stabilizes depth token emission. A soft depth reconstruction technique enables fully end-to-end differentiable depth training.\footnote{Codebase will be released upon publication.}
    
    \item \textbf{Perception-enhanced data.} We curate a 56K-example joint dataset that pairs segmentation masks and depth priors with language supervision, augmenting RefCOCO/+/g with aligned depth tokens and attribute descriptions to steer intermediate reasoning. \footnote{Dataset will be released upon publication.}
    
    \item \textbf{State-of-the-Art Performance.} Perceptio achieves SOTA on all three referring segmentation benchmarks (RefCOCO/+/g), a +10.3\% improvement on HardBLINK spatial reasoning, and a +1.0\% gain on MMBench general VQA, demonstrating that explicit in-sequence perception materially strengthens spatial grounding across diverse tasks.
    
\end{enumerate}


\begin{figure}
\vspace{-20pt}
    \centering
    \includegraphics[width=0.99\linewidth]{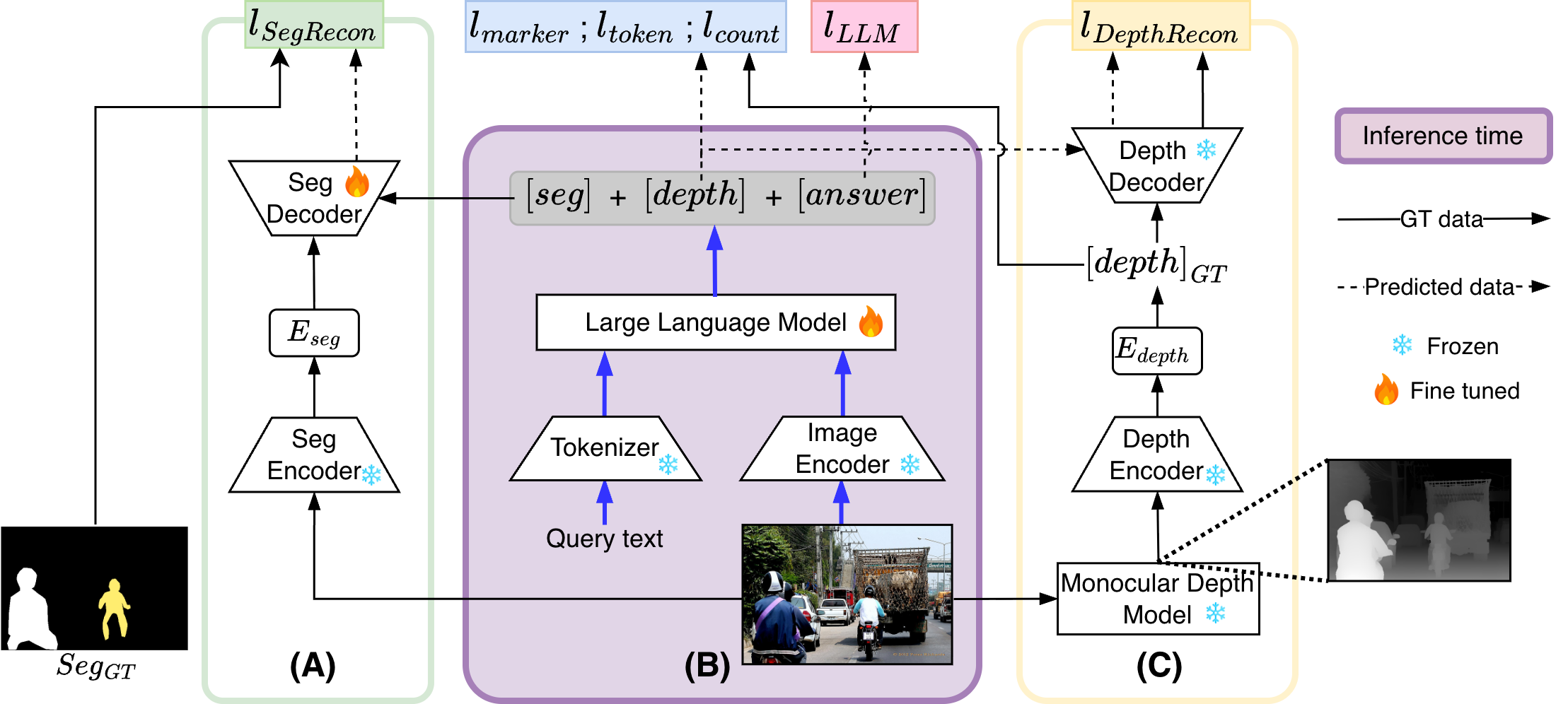}
    \caption{The \modelname model. The language model takes text queries and image features as input to generate the desired text sequence \textbf{(B)}. During training time, segmentation \textbf{(A)} and depth \textbf{(C)} teacher models supervise (via loss functions) the LVLM to accurately generate the intermediate perception tokens and the answer text output tokens.}
    \label{fig:main-model-arch}
\vspace{-20pt}
\end{figure}
\vspace{-16pt}



\section{Related Work}
\label{sec:related_work}
\vspace{-8pt}

\subsection{Large Vision-Language Models (LVLMs)}
Recently, LVLMs have demonstrated remarkable progress. These models integrate tokenized visual features with language tokens, feeding the combined representation into a pre-trained Large Language Model (LLM) to understand and generate responses that span both visual and linguistic domains \cite{2023visionllm,zhu2023minigpt,Chen2023MiniGPTv2LL}. The latest landscape of LVLMs, including robust architectures like LLaVA\cite{liu2023llava}, GPT-4v\cite{gpt4v2023}, and their contemporaries, has pushed the boundaries of general-purpose visual reasoning, complex dialogue, and detailed image captioning. However, a critical review shows these models are better at semantic understanding (i.e., knowing what is in an image) than spatial understanding (i.e., knowing where things are), because their architectures are not explicitly designed to model spatial awareness.

Rather than being explicitly modeled, complex spatial relationships such as relative and absolute positions are typically assumed to emerge from training at scale; as a result, spatial reasoning is rarely treated as a first‑class, foundational objective. For example, despite its scale, InternVL2.5‑26B achieves only 33.1\% average accuracy on HardBlink's ``closer‑to‑camera'' point‑selection task (details in Table \ref{tab:relative_depth}). This underscores that spatial understanding remains a notable weakness in multi-modal LLMs and does not reliably emerge from scale alone.

\vspace{-8pt}
\subsection{Perception Guidance in LVLMs}

Despite rapid progress in LVLMs, fine‑grained grounding and spatial reasoning remain difficult because text decoders often infer geometry from pooled features without explicit spatial cues. Two‑stage pipelines, such as, LLM controllers wrapped around LISA \cite{lai2023lisa} improve segmentation, as do token emitting LVLM variants, but they externalize perception and rarely feed masks back into the reasoning loop \cite{lai2023lisa,kirillov2023segment,xia2023gsva}. PerceptionGPT \cite{Pi2023PerceptionGPTEF} brings perception into the sequence by learning a dynamic token that encodes boxes and masks, boosting performances on Referring Expression Segmentation (RES), yet, remains limited to 2D semantics \cite{Mao2015GenerationAC}. Sa2VA further unifies an LLM with SAM2 to produce query‑grounded masks for images and videos, advancing RES while still operating on planar cues \cite{yuan2025sa2vamarryingsam2llava}. In parallel, AURORA introduces "perception tokens" that discretize mid‑level signals, most notably monocular depth via a VQ‑VAE codebook yielding sizable gains on depth and counting; however, it neither outputs segmentation masks for grounding nor fuses 2D semantics with 3D geometry in one model, and it can degrade general VQA performance \cite{Bigverdi2024PerceptionTE}. 

Evidence from DenseWorld‑1M shows that leading LVLMs still miss small objects and misalign references, underscoring inadequate spatial grounding \cite{li2025denseworld1mdetaileddensegrounded}. These limitations stem from LVLMs natively emitting text rather than dense maps. Injecting intermediate 2D and 3D cues helps, but purely text-decoder LVLMs still under perform at spatial understanding. Simultaneously, specialist pipelines excel on targeted spatial tasks yet trade-off broad conversational ability. Similarly, metric-depth only approaches (e.g., DepthLM \cite{cai2025depthlmmetricdepthvision}) do not unify 2D semantics with 3D geometry. To our knowledge, no prior work jointly optimizes complementary objectives for 2D semantic segmentation and 3D depth reasoning within a single LVLM. \textbf{\modelname} closes this gap by injecting SAM2‑based semantic segmentation tokens and discretized depth tokens into the sequence, enabling explicit spatial reasoning and yielding state‑of‑the‑art (SOTA) grounding performance on multiple tasks.

\vspace{-10pt}

\section{Methods}
\label{sec:methods}

\vspace{-8pt}
We introduce \modelname, a perception-enhanced LVLM that explicitly incorporates visual segmentation and depth cues into its generation process. In this section, we first describe the model architecture and the insertion of semantic segmentation and discretized depth tokens into the autoregressive sequence (\ref{sec:methods-model-architecture}). Then, we detail the procedure for generating perception tokens and explain how the model learns from our perception conditioned generation pattern (\ref{sec:methods-perception-guidance}). Next, we describe the model’s inference‑time behavior (\ref{sec:methods-model-inference}), followed by the multi‑task objective (\ref{sec:method-loss-function}) and experimental setup (\ref{sec:methods-experimental-setup}).

\vspace{-8pt}
\subsection{Model Architecture}
\label{sec:methods-model-architecture}
Figure~\ref{fig:main-model-arch} provides an overview of our approach: \modelname. Given an input image and a text query, the system routes visual signals through three complementary pathways: (i) a standard image encoder for semantic appearance features; (ii) a \emph{frozen} SAM encoder for segmentation-aware representations; and (iii) a \emph{frozen} pre-trained depth Quantized Variational Autoencoder (VQ-VAE) codebook that discretizes image depth. The core LLM consumes the encoded image features together with the query and produces an autoregressive sequence that interleaves natural-language tokens with perception-control tokens. In particular, it predicts special $\texttt{[seg]}$ token to request segmentation and a sequence of discrete depth tokens $\texttt{[depth]}$ to represent depth. These tokens trigger task-specific decoders: when \texttt{[seg]} appears, a SAM2 decoder reconstructs segmentation masks; when \texttt{[depth]} appears, a depth decoder maps the discrete codes back to a continuous depth map via the VQ-VAE codebook. During training, we fine-tune the SAM2 decoder to learn the special segmentation tokens, supervising it with reconstruction losses against ground-truth masks. In contrast, the depth branch (codebook and decoder) is kept frozen: the LVLM is trained only to generate depth tokens that index the pre-trained codebook, enabling depth reconstruction without updating the depth decoder. This unified design enables \modelname to perform language generation, referring expression segmentation, and depth reasoning within a single autoregressive framework, making perception a first-class part of the language-modeling objective rather than a post-hoc step.

\vspace{-10pt}
\subsection{Model Learning}
\vspace{-8pt}
\label{sec:methods-perception-guidance} 
\paragraph{Perception Enhanced Generation.}
Perception enhanced generation refers to our strategy of guiding the LVLM’s generation with intermediate visual cues. We enforce a specific output format: the model’s generated token sequence must contain a segmentation token block and a depth token block before the final textual answer. Formally, the sequence is structured as:

\vspace{-14pt}
\begin{align}
\label{eq:tokens}
[\texttt{seg}]\,[d_{\text{start}}, d_1, d_2, \ldots, d_n, d_{\text{end}}]\,[t_1, t_2,\ldots, t_m]
\end{align}

where $\left[\texttt{seg}\right]$ is a special control token whose embedding conditions the segmentation decoder to output a query-grounded mask, \([d_{\text{start}}, d_1, \ldots, d_n, d_{\text{end}}]\) are the discretized depth tokens, and \([t_1, t_2,\ldots, t_m]\) represents the text answer tokens. The model is trained to always emit these in order: first the segmentation, then the depth, and then the answer. The motivation for this enforced ordering arises from the autoregressive nature of the decoder—by generating perceptual tokens first, the model effectively performs a chain-of-thought reasoning based on the scene’s spatial structure before formulating a final answer. This approach injects explicit spatial awareness into the language model's output by requiring the model to explicitly generate its visual perception in the form of segmentation masks and depth maps before producing the final response to the query.

\paragraph{Depth Codebook.}
To capture fine-grained 3D structure, inspired by \cite{ning2023tokensunifyingoutputspace}, we construct a depth codebook using a VQ-VAE with codebook size $K$ \cite{oord2018neuraldiscreterepresentationlearning}. We first obtain reliable continuous depth maps with the depth-specialist model Depth Anything V2 \cite{yang2024depth}, then discretize them into depth tokens via vector quantization, enabling seamless integration into our token-based framework. In contrast to prior work that learns a codebook on a single, specialized depth dataset \cite{ning2023tokensunifyingoutputspace}, we train on all depth maps derived from the same scene-image corpora (\ref{sec:methods-experimental-setup}) used to finetune the LLM. This distributional alignment improves robustness and strengthens depth perception. The resulting VQ-VAE depth codebook serves as a broadly generalizable prior and an augmentation signal that guides the LLM to generate accurate depth tokens.

In this setup, each depth map is encoded as a grid of embeddings, where the nearest-neighbor distance is used to identify the closest entry in the codebook. The VQ-VAE decoder reconstructs the depth map from the sequence of latent codes, and the entire model is trained with mean-squared-error (MSE) reconstruction loss to ensure accurate reconstruction. During inference, we patchify the depth map into a \(\sqrt{n}\times\sqrt{n}\) grid of code indices, resulting in an \(n\)-token sequence where each token represents one of the $K$ discrete depth values in the depth codebook, labeled \(d_1\) to \(d_{n}\). The depth token sequence starts with a special $[d_{start}]$ token and ends with a special $[d_{end}]$ token, adding a total of \(K+2\) depth-related tokens (\(K\) depth values plus two special tokens) to the model's vocabulary.

Concretely, for each training sample we augment the textual prompt so that the target sequence first includes perception tokens (segmentation and depth) followed by the textual answer. Exposure to these augmented training
instances encourages the model to condition its reasoning on explicit
segmentation and depth cues. At inference, these perception tokens are not produced by arbitrary prompts; we use lightweight prompt templates with special tokens to reliably elicit the intermediate segmentation and depth tokens alongside the final answer.  Our perception enhanced design helps the model to internalize these perceptual cues and demonstrates improved performance on tasks requiring fine‑grained grounding.

\vspace{-8pt}
\subsection{Model Inference}
\label{sec:methods-model-inference}
\vspace{-8pt}

At test time, given an input image \(I\) and textual prompt \(q\), we tokenize \(q\) and encode \(I\) into visual tokens. The text and image tokens are concatenated and fed to the LVLM, which autoregressively emits an interleaved sequence of control and content tokens same as defined in Eq. (\ref{eq:tokens}). Each group gates a downstream prediction head, specifically:

\noindent\textit{Segmentation Head:} Emitting \texttt{[seg]} activates the SAM2 decoder, which fuses the \texttt{[seg]} query from the LLM with dense features from the SAM2 encoder to predict a segmentation mask \(\hat{\mathbf{M}}\). The mask type (e.g., referring, instance, or semantic) is determined by the task implied by the prompt.

\noindent\textit{Depth Head:} The depth subsequence \(d_{1:n}\) is interpreted as indices into a VQ--VAE codebook and decoded to reconstruct a dense depth map \(\hat{\mathbf{D}}\).

\noindent\textit{Text Head:} The text subsequence \(\mathbf{t}\) is detokenized to form the natural-language response.

This design unifies language, segmentation, and depth outputs within a single coherent token sequence.
\textit{It is important to note that}, during inference the generated \texttt{[seg]} and \texttt{[depth]} tokens are available to create 2D and 3D grounding visualizations via their respective teacher models. However, the trained LVLM model operates independent of the teacher branches to generate the desired text response for downstream tasks.

\subsection{Loss Functions}
\label{sec:method-loss-function}

Effective spatial reasoning requires carefully designed supervision signals. To this end, we design novel loss functions for 3D depth information generation (\ref{subsec:3d-loss}), while leveraging the standard LLM loss (\ref{subsec:llm-loss}) for text generation and segmentation loss (\ref{subsec:2d-loss}) for the 2D segmentation feedback, respectively. We optimize all tasks in a single fine-tuning stage by minimizing the total loss defined as follows.
\begin{equation}
\label{eq:total}
L_{\text{total}} \;=\; L_{\text{LLM}} \;+\; L_{\text{SegRecon}} \;+\; \lambda_d\,L_{\text{depth}} \;+\; \lambda_r\,L_{\text{DepthRecon}}
\end{equation}
where, $\lambda_d$ \& $\lambda_r$ are weights for respective loss contributions. Next, we explain each loss term in detail.

\subsubsection{LLM Loss}
\label{subsec:llm-loss}
LLM loss is the standard teacher‑forced next‑token negative log‑likelihood for the decoder conditioned on image features: \(L_{\text{LLM}}=-\frac{1}{T}\sum_{t=1}^{T}\log p_{\theta}(y_t \mid y_{<t}, \phi(\mathcal{I}))\).

\subsubsection{2D Supervision} 
\label{subsec:2d-loss}
Here, we aim for the LLM generated \texttt{[seg]} token to improve, such that, it creates accurate segmentation masks in the segmentation decoder. We use a reconstruction loss as 2D supervision between the ground truth segmentation mask $SEG_{GT}$ and the reconstructed segmentation mask using the generated \texttt{[seg]} token. We combine pixel-wise cross-entropy and DICE loss:
\begin{equation}
\label{eq:seg}
L_{\text{SegRecon}} \;=\; L_{\text{CE}} \;+\; L_{\text{D}}~.
\end{equation}

\subsubsection{3D Supervision}
\label{subsec:3d-loss}
Depth supervision comprises (i) a depth \emph{token generation} loss ($L_{\text{depth}}$), and (ii) a differentiable \emph{soft reconstruction} loss ($L_{\text{DepthRecon}}$). 



\paragraph{Depth Token Generation Loss.}
We fine-tune the LLM with LoRA to incorporate depth information by adding special depth tokens to its vocabulary. However, relying solely on the standard next-token cross-entropy loss may not be sufficient to ensure these tokens are generated as intended. To better ground the model in the meaning and proper use of depth tokens, we introduce additional regularization terms. Specifically, we propose a suite of novel loss functions targeted at encouraging accurate and consistent depth-token generation. Depth is emitted as a bracketed sequence 

\[
[\,d_{\text{start}},\, d_1,\dots,d_n,\, d_{\text{end}}\,]
\]

with \(n\) tokens from a VQ-VAE codebook.
For each sample \(b\), let \((s_b,e_b)\) be the start/end indices with 
\(\mathbf{y}_{b,s_b}=d_{\text{start}}\) and 
\(\mathbf{y}_{b,e_b}=d_{\text{end}}\).
Define the interior length
\[
l_b \;=\;
\begin{cases}
e_b - s_b - 1, & \text{if a valid span exists,}\\
0,              & \text{otherwise.}
\end{cases}
\]
The depth token generation loss is a composite loss to align \emph{when} the span begins/ends ($L_{\text{marker}}$), \emph{what} codes fill it ($L_{\text{token}}$), and \emph{how many} are produced ($L_{\text{count}}$).
\begin{equation}
\label{eq:depth_total}
L_{\text{depth}}
\;=\;
\lambda_m\,L_{\text{marker}}
\;+\;
\lambda_t\,L_{\text{token}}
\;+\;
\lambda_c\,L_{\text{count}}
\end{equation}

(Values of the coefficients reported in \ref{sec:methods-experimental-setup})

\noindent\textbf{Marker Loss.}
To ensure depth start token $d_{start}$ and depth end token $d_{end}$ are generated at correct positions, we propose a marker loss:
\begin{equation}\label{eq:marker_loss}
L_{\text{marker}}=\frac{1}{B}\sum_{b=1}^{B}\Big(\mathbf{1}_{\{s_b\neq\emptyset\}}\mathrm{CE}(\mathbf{z}_{b,s_b-1},\mathbf{y}_{b,s_b})+\mathbf{1}_{\{e_b\neq\emptyset\}}\mathrm{CE}(\mathbf{z}_{b,e_b-1},\mathbf{y}_{b,e_b})\Big).
\end{equation}
where $B$ is the batch size, $CE(\cdot, \cdot)$ is token-level cross-entropy, $\mathbf{z}\!\in\!\mathbb{R}^{B\times T\times V}$ are decoder logits,
$\mathbf{y}\!\in\!\mathbb{N}^{B\times T}$ are ground-truth tokens. $T$ is the sequence length and $V$ is the vocabulary size. The indicator $\mathbf{1}_{\{s_b \neq \emptyset\}}$ equals 1 when a valid depth span is found in sample $b$ (i.e., both $d_{\text{start}}$ and $d_{\text{end}}$ are present), and 0 otherwise.

\noindent\textbf{Token Loss.}
To ensure correct depth token values are generated by LLM, we proposed a token loss defined as:
\begin{equation}
\label{eq:token_loss}
L_{\text{token}}
=
\frac{1}{B}
\sum_{b=1}^{B}
\frac{\mathbf{1}_{\{l_b>0\}}}{\max(l_b,1)}
\sum_{t=s_b+1}^{e_b-1}
\mathrm{CE}\!\big(\mathbf{z}_{b,t-1},\,\mathbf{y}_{b,t}\big).
\end{equation}

\noindent\textbf{Count Loss.}
Last, to encourage the LLM to produce sequences of the desired length \(n\), we proposed a penalty term that activates when the generated length deviates from  \(n\):
\begin{equation}
\label{eq:count_loss}
L_{\text{count}}
=
\frac{1}{B}
\sum_{b=1}^{B}
\log\!\big(1 + \lvert l_b - n\rvert\big).
\end{equation}

\paragraph{Soft Depth Reconstruction.}
To decode depth in a differentiable manner, inspired by \cite{ning2023tokensunifyingoutputspace}, we replace hard codeword selection with a soft-merging technique of codebook embeddings. The model predicts a probability distribution over the codebook and we form a soft token by weighting each embedding with its predicted probability. This continuous relaxation maps discrete tokens into a smooth embedding space, allowing gradients from the depth reconstruction objective to flow through the tokenization stage and enabling fully end-to-end training.
For each timestep \(t\) inside the depth span, restrict logits to depth-code index set \(\mathcal{D}\) and compute
\begin{equation}
p_t(k) \;=\; 
\frac{\exp\!\big(\mathbf{z}_{t,k}\big)}{\sum_{j\in\mathcal{D}}\exp\!\big(\mathbf{z}_{t,j}\big)}\,,
\quad k\in\mathcal{D},
\end{equation}
then form the expected latent
\begin{equation}
\tilde{\mathbf{z}}_t \;=\; \sum_{k \in \mathcal{D}} p_t(k)\,\mathbf{e}_k\,.
\end{equation}
where $\mathbf{e}_k$ is the VQ-VAE codebook vector for index $k\!\in\!\mathcal{D}$ and $p_t$ is the softmax over depth codes at step $t$. 
The sequence \(\{\tilde{\mathbf{z}}_t\}\) is truncated to \(n\), reshaped to a \(\sqrt{n}\times\sqrt{n}\) grid, and decoded by the VQ-VAE into a predicted depth map \(\hat{\mathbf{Y}}\). 
We minimize
\begin{equation}
\label{eq:recon}
L_{\text{DepthRecon}} \;=\; \frac{1}{B}\sum_{b=1}^{B} \big\| \hat{\mathbf{Y}}_b - \mathbf{Y}_b \big\|_2^2\,.
\end{equation}

\subsection{Experimental Setup}
\label{sec:methods-experimental-setup}

\noindent
\textbf{Dataset Curation:}
We build a joint dataset by augmenting RefCOCO, RefCOCO+, and RefCOCOg \cite{yu2016modeling,Mao2015GenerationAC}, referring expression segmentation benchmarks where each example pairs a free‑form phrase with the pixel‑accurate mask of the mentioned object—with complementary supervision for depth and description. Concretely, for every referring expression we (i) convert the ground‑truth mask into a compact sequence of segmentation tokens; (ii) attach aligned depth tokens that encode the quantized depth of the same region; and (iii) add a concise, attribute‑focused one‑sentence object description. All signals are unified in a single instruction–output format so the model learns, from one prompt, to ground the phrase, emit the mask (via seg tokens), infer scene layout (via depth tokens), and verbalize salient attributes. We retain official splits, perform consistency and ambiguity filtering, and deduplicate near‑overlapping samples to preserve quality. This multi‑signal curation turns classic referring data into a scalable corpus for joint perception and reasoning.

\noindent
\textbf{Training Dataset:} We train \textbf{\modelname} using only image-based corpora covering (i) image question and answering and image chat, (ii) image-level text-driven segmentation, (iii) depth guided chain-of-thought data, and (iv) joint dataset for segmentation, depth, and text. Our training set comprises approximately $1.1M$ image--text pairs drawn from three sources: (i) 665K LLaVA‑1.5 instruction-tuning samples for image QA and chat \cite{Liu2023LLaVAPlusLT}, (ii) 214K grounding conversation generation samples for image-level text-driven segmentation \cite{hanoona2023GLaMM}, and (iii) we use our curated datasets with synthetic and unique 60K ADE20k with Perception tokens dataset, inspired by \cite{Bigverdi2024PerceptionTE}. We use the referring-expression segmentation datasets—RefCOCO (17K), RefCOCO+ (17K), and RefCOCOg (22K)—all built on MS COCO 2014 images. From these resources, we also curate a joint dataset with referring-expression segmentation by augmenting captions with depth tokens, yielding a total of 56K examples.
Because InternVL2.5 \cite{chen2024expanding} is already pre-trained on large-scale image QA data, we fine-tune with the LLaVA‑1.5 corpus to preserve QA capability while adapting the model to grounding and segmentation.

\noindent
\textbf{Evaluation:} For the grounding task, we evaluate RefCOCO, RefCOCO+ and RefCOCOg on validation datasets;  To assess broader vision-language capabilities, we also employ the recent MME \cite{fu2023mme}, MMBench \cite{liu2023mmbench}, and SEED-Bench \cite{li2023seed} benchmarks, which cover a wide range of multi-modal tasks including visual question answering, captioning, and reasoning and report performance via objective metrics such as multiple-choice accuracy. For science and diagram VQA evaluation, we test on AI2D \cite{kembhavi2016diagram}, MMStar \cite{chen2024we}, and the ScienceQA \cite{lu2022sqa} collectively covering diagram understanding and grounded reasoning. Additionally, we include the HardBLINK variants with 3, 4, and 5 marked points per image \cite{Bigverdi2024PerceptionTE}. These tasks challenge the model's spatial understanding by requiring it to identify relationships like which point is closest, and we measure success as the percentage of queries answered correctly. (relative depth accuracy)

\noindent
\textbf{Metrics:} For image referring segmentation, we evaluate by measuring the alignment between the predicted mask and ground truth, using Intersection-over-Union (IoU)–based metrics (with an IoU $>$ 0.5 defining a correct prediction). For image QA and image chat, we follow prior work and report the standard benchmark-specific metrics used by those works \cite{fu2023mme,li2023seed}. For spatial reasoning, we report the accuracy of correctly answered multiple-choice questions.

\noindent
\textbf{Implementation Details:}
We train our model on 64 NVIDIA A100 GPUs for approximately 24 hours. Following InternVL design, we adopt a maximum sequence length of 8,192 tokens. The model is optimized using AdamW with a learning rate of $4\times 10^{-5}$. We employ a batch size of 1 per device with gradient accumulation over 8 steps, resulting in an effective batch size of 512. For LoRA parameters, we set the rank to 256, chosen to provide sufficient capacity for the model to learn the new depth and segmentation token embeddings alongside language adaptation. The learning rate schedule consists of a linear warmup for the first 5\% of training, followed by cosine annealing decay to zero. Gradient clipping is applied with a maximum norm of 1.0. For the depth VQ-VAE, we use a codebook dimension of $K$ = 128. The LLM generates $n$=100 depth tokens. The segmentation loss combines cross-entropy loss (weight=1.0) and DICE loss (weight=0.25), both with sigmoid activation. The VQ-VAE reconstruction loss is weighted at 1.0. During evaluation, we distribute inference across 8 GPUs to handle the long-context processing efficiently. In our experiments, we set the loss weights to \(\lambda_m = 0.3\) (marker), \(\lambda_t = 0.5\) (token), \(\lambda_c = 0.2\) (count), \(\lambda_d = 1.0\) (depth), and \(\lambda_r = 1.0\) (depth reconstruction).
We train both 8B and 4B variants of our model for our main results, and ablation experiments use the 8B variant.

\noindent
\textbf{Inference Cost:}
Despite generating additional perception tokens, Perceptio incurs negligible inference overhead. For dense caption generation, Perceptio-8B takes 3.52 seconds per 100 tokens compared to 3.53 seconds for Sa2VA-8B, with comparable FLOPs (4.06T vs.\ 4.66T). At inference time, only the LVLM is required; the teacher models (SAM2 encoder, Depth Anything V2) are not needed unless the application explicitly requires segmentation mask or depth map outputs.

\section{Results}
\label{sec:results}
\vspace{-5pt}

\begin{table}[t!]
    \centering
    \caption{\small{Merged results across image referring segmentation, image chat, and image-level benchmarks. RefCOCO/+/g report cIoU. For MME, we list Perception (P), Cognition (C), and Total (T{=}P{+}C). ``--'' indicates not reported.}}
    \resizebox{\textwidth}{!}{
    \begin{tabular}{c|ccc|ccccc|ccc}
    \toprule[0.2em]
    \multirow{2}{*}{Method} 
      & \multicolumn{3}{c|}{Image Segmentation} 
      & \multicolumn{5}{c|}{Image Chat / Understanding} 
      & \multicolumn{3}{c}{Diagram / Science QA} \\
    ~ 
      & \scriptsize{RefCOCO~\cite{kazemzadeh2014referitgame}} 
      & \scriptsize{RefCOCO+~\cite{kazemzadeh2014referitgame}} 
      & \scriptsize{RefCOCOg~\cite{yu2016modeling}}
      & \scriptsize{MME-P~\cite{fu2023mme}} 
      & \scriptsize{MME-C~\cite{fu2023mme}} 
      & \scriptsize{MME-T} 
      & \scriptsize{MMBench~\cite{liu2023mmbench}} 
      & \scriptsize{SEED-Bench~\cite{li2023seed}}
      & \scriptsize{AI2D~\cite{kembhavi2016diagram}} 
      & \scriptsize{MMStar~\cite{chen2024we}} 
      & \scriptsize{SQA$^{\rm test}$~\cite{lu2022sqa}} \\
    \midrule
    LLAVA-1.5-13B~\cite{liu2023improvedllava} & -- & -- & -- & -- & -- & 1531 & 68.8 & 70.1 & -- & -- & -- \\
    Video-LLaVA-7B~\cite{lin2023videollava}   & -- & -- & -- & -- & -- & --   & 60.9 & --   & --  & -- & -- \\
    mPLUG-Owl3-8B~\cite{ye2024mplug}          & -- & -- & -- & -- & -- & --   & 77.6 & --   & -- & -- & -- \\
    InternVL2-8B~\cite{chen2024far}           & -- & -- & -- & -- & -- & --   & 81.7 & \textbf{76.2} & \textbf{83.8} & -- & -- \\
    PixelLM-7B~\cite{ren2023pixellm}          & 73.0 & 66.3 & 69.3 & 309 & 135 & 444  & 17.4 & --   & 0.0 & -- & -- \\
    LaSagnA~\cite{wei2024lasagna}             & 76.8 & 66.4 & 70.6 & 0   & 0   & 0    & 0.0  & --   & 0.0 & -- & -- \\
    GLaMM-7B~\cite{hanoona2023GLaMM}          & 79.5 & 72.6 & 74.2 & 14  & 9   & 23   & 36.8 & --   & 28.2 & -- & -- \\
    LLaVA-G-7B~\cite{zhang2025llavagrounding} & 77.1 & 68.8 & 71.5 & --  & --  & --   & --   & --   & -- & -- & -- \\
    GSVA-13B~\cite{xia2023gsva}               & 79.2 & 70.3 & 75.7 & --  & --  & --   & --   & --   & -- & -- & -- \\
    OMG-LLaVA-7B~\cite{OMGLLaVA}              & 78.0 & 69.1 & 72.9 & 1177& 235 & 1412 & 47.9 & 56.5 & 42.9 & -- & -- \\
    VISA-13B~\cite{yan2024visa}               & 72.4 & 59.8 & 65.5 & --  & --  & --   & --   & --   & -- & -- & -- \\
    Sa2VA-4B~\cite{yuan2025sa2vamarryingsam2llava} & 80.4 & 74.3 & 76.7 & 1553 & 540 & 2093 & 76.8 & 72.6 & 79.9 & 53.7 & 95.8 \\
    Sa2VA-8B~\cite{yuan2025sa2vamarryingsam2llava} & 81.9 & 76.5 & 78.9 & 1651 & 578 & 2229 & 82.4 & 75.5 & 82.1 & 60.3 & 96.8 \\
    \midrule
    \modelname-4B (Ours) & 81.7 & 76.9 & 78.9 & \textbf{1710} & 615 & \textbf{2325} & 82.0 & 74.4 & 81.5 & 57.7 & 97.2 \\
    \modelname-8B (Ours) & \textbf{82.7} & \textbf{77.9} & \textbf{80.0} & 1654 & \textbf{628} & 2282 & \textbf{83.4} & 75.7 & 83.4 & \textbf{64.2} & \textbf{98.3} \\
    \bottomrule[0.1em]
    \end{tabular}
    }
    \label{tab:merged_benchmarks}
\end{table}


\subsection{Main Results}
\label{sec:results-implementation}

\noindent \textbf{\emph{Quantitatively}}, we evaluate \modelname across a comprehensive suite of benchmarks spanning referring image segmentation, multimodal dialogue, and relative depth reasoning. As summarized in Table~\ref{tab:merged_benchmarks}, \modelname-8B sets a new state of the art on all three referring segmentation datasets—82.7\% on RefCOCO, 77.9\% on RefCOCO+, and 80.0\% on RefCOCOg—surpassing the best prior Sa2VA-8B by +1.1/+1.7/+1.3 points  in cIoU score respectively. On image chat evaluations, \modelname-8B achieves the strongest MME perception/cognition scores (1654/628) and the best MMBench accuracy (83.4), while remaining highly competitive on SEED-Bench (75.7, within 0.5 of InternVL2-8B). The lighter \modelname-4B mirrors these trends, already outperforming larger baselines (e.g., 81.7/76.9/78.9 on RefCOCO/+/g and 1710/615 on MME). On the HardBLINK relative depth task (Table~\ref{tab:relative_depth}), our depth tokens and intermediate reasoning yield substantial gains in accuracy: \modelname-8B attains 75.8/71.0/66.1 for 3/4/5 points with a 71.0 average, improving over LLaVA-Aurora by +8.9/+10.5/+11.3 and +10.3 points on average. The accompanying figure provides a visual comparison of these results; additional qualitative examples across models and datasets are included in the supplementary materials.

\begin{wraptable}[11]{r}{0.52\columnwidth}
\vspace{-35pt}
\centering
\caption{\small{Relative depth accuracy on HardBLINK. Our Perceptio-4B/8B—using depth tokens and intermediate reasoning—outperform prior baselines.}}
\label{tab:relative_depth}
\resizebox{\linewidth}{!}{
\begin{tabular}{c|ccc|c}
\toprule[0.2em]
\textbf{Model} & \multicolumn{4}{c}{\textbf{HardBLINK}} \\
~ & 3 Points & 4 Points & 5 Points & Average \\
\midrule
Sa2VA-8B~\cite{yuan2025sa2vamarryingsam2llava} & 21.0 & 17.7 & 25.0 & 21.2 \\
InternVL2.5-26B~\cite{chen2024expanding} & 41.1 & 31.5 & 26.6 & 33.1 \\
LLaVA 1.5 13B~\cite{liu2023improvedllava} & 35.5 & 37.9 & 29.0 & 34.1 \\
Fine-tuned LLaVA~\cite{Bigverdi2024PerceptionTE} & 58.9 & 52.4 & 41.1 & 50.8 \\
LLaVA-Aurora~\cite{Bigverdi2024PerceptionTE} & 66.9 & 60.5 & 54.8 & 60.7 \\
\midrule
\modelname-4B (Ours) & 69.4 & 66.9 & 59.7 & 65.3 \\
\modelname-8B (Ours) & \textbf{75.8} & \textbf{71.0} & \textbf{66.1} & \textbf{71.0} \\
\bottomrule[0.1em]
\end{tabular}
}
\vspace{-0.6\baselineskip}
\end{wraptable}

\begin{figure}
    \centering
    \vspace{-10pt}
    \includegraphics[trim={2.8cm 0 0 0}, clip, width=0.85\linewidth]{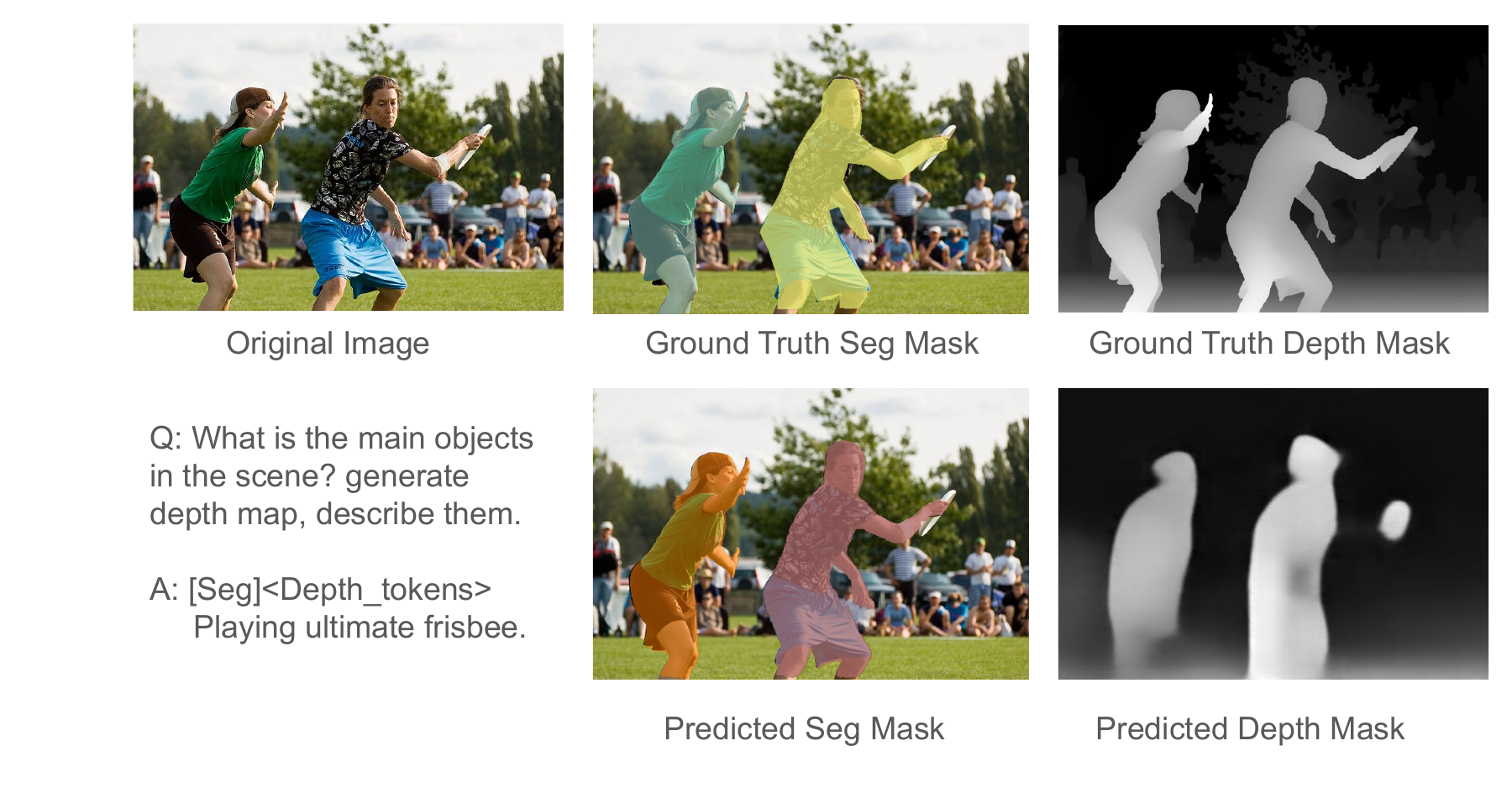}
    \vspace{-10pt}
    \caption{Comparison between ground-truth and predicted outputs for a scene depicting two players engaged in an ultimate frisbee game. The first row shows the original image, ground-truth segmentation, and depth masks. The second row displays the question and the corresponding model predictions for text, segmentation mask, and depth map, demonstrating accurate recognition of the main objects and spatial depth relationships. }
    \vspace{-20pt}
    \label{fig:visualization}
\end{figure}

\begin{figure}
    \centering
    \includegraphics[width=\linewidth]{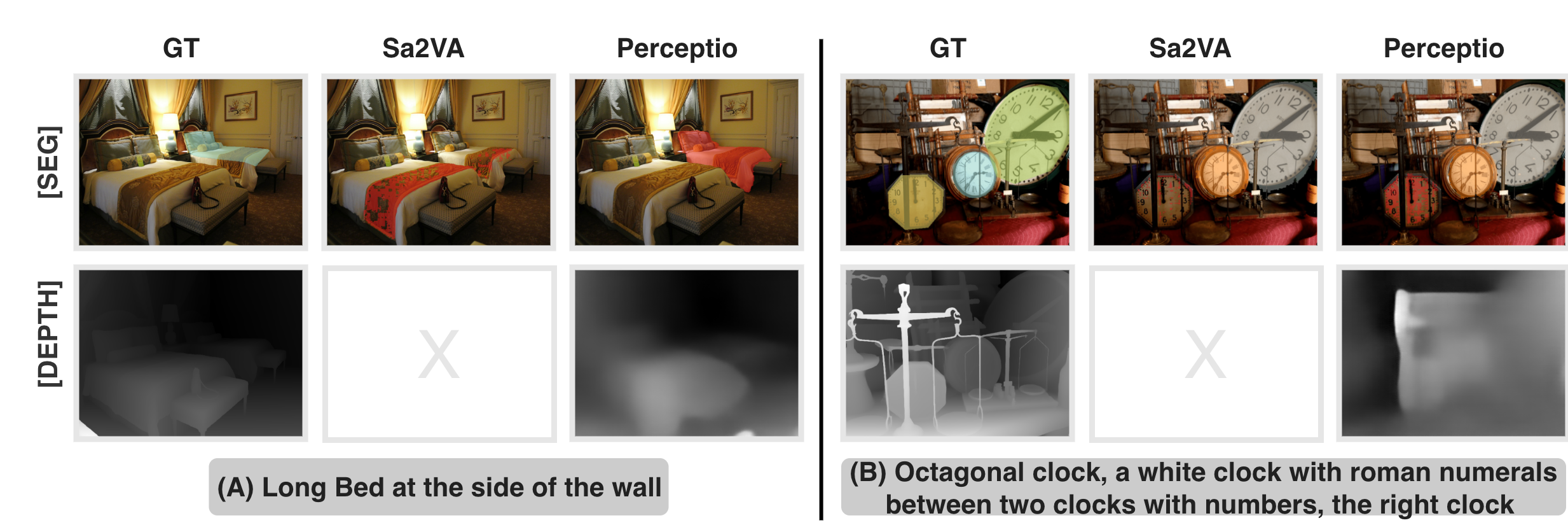}

    \caption{Perceptio evaluated on RefCOCO qualitative results. We compare our model with Sa2VA on a referring expression. Instance masks are colorized for visibility; depth maps are shown in grayscale (lighter = nearer). Our predictions align with semantic boundaries better than Sa2VA by capturing the expected depth layering across the image. Note: "×" denotes that no depth perception in Sa2VA.}
    \label{fig:sa2va-vs-perceptio}
\end{figure}

\begin{wrapfigure}{r}{0.45\linewidth}
  \vspace{-100pt}
  \centering
  \includegraphics[width=\linewidth]{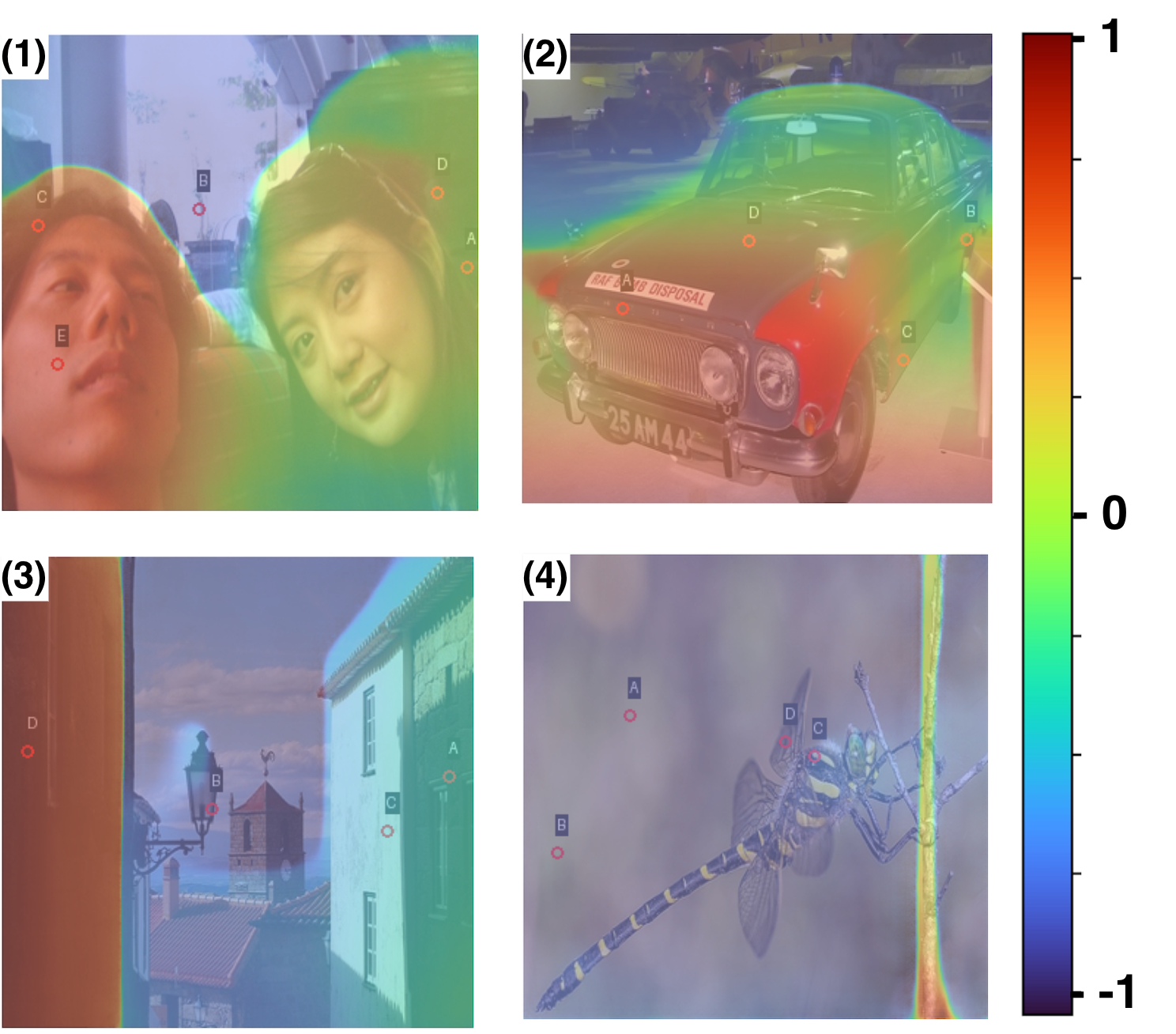}
  \caption{Perceptio (Ours) predictions on HardBLINK dataset with overlay of color depth maps. Correct predictions in samples 1,2,3 and incorrect prediction in sample 4.}
  \vspace{-15pt}
  \label{fig:hardblink}
\end{wrapfigure}

\noindent \textbf{\emph{Qualitatively}}, we visualize samples in the RefCOCO dataset to internalize the impact of joint perception. In Figure \ref{fig:visualization} we see the reconstructed depth maps, and predicted segmentation masks along with the question and the generated answer. We see clear depth separation in the 3D depth map, along with semantic segmentation boundaries of the object. In Figure \ref{fig:sa2va-vs-perceptio}, we see samples where Sa2VA model fails to predict the correct segmentation masks, whereas, Perceptio model generates accurate depth maps and corresponding segmentation masks for the same input queries. 
Further highlighting the importance of depth perception, in Figure \ref{fig:hardblink}, \modelname make correct predictions where depth maps capture 3D information, and make error in sample 4 where depth map marks all objects as background.

\vspace{-5pt}

\subsection{Ablation Studies}
\label{sec:results-ablation}



\subsubsection{Impact of Perception}
We introduce joint 2D and 3D perception ability in LVLMs. A natural question emerges on, \textit{how significant is each source of perception?} In this ablation, we train the model with 1) No depth signal, 2) No segmentation signal and compare it with \modelname. For each variant, we remove the underlying dataset, task and loss function for the respective perception signal.

\textbf{2D Only Perception:} As shown in Table \ref{tab:hardBLINK}, the performance on the depth reasoning task of HardBLINK significantly decreases (the average accuracy drops from 71.0\% to 45.2\%, a decline of 25.8 percentage points) when the depth tokens and the depth branch of \modelname are removed. Notably, general VQA metrics slightly \emph{improve} without depth tokens (Table~\ref{tab:compact_results}), indicating a mild optimization tension between depth token generation and text-only tasks. However, the substantial HardBLINK collapse demonstrates that depth tokens are essential for 3D spatial reasoning, the core capability targeted by our work. We discuss this trade-off further in Sec.~5.

\textbf{3D Only Perception:}
We evaluate a depth-only variant that removes the segmentation tokens while retaining depth tokens (Table~\ref{table:no-seg-ablation}). This \emph{3D-only} setting probes, whether the model can rely purely on geometric cues to structure the scene. As shown in Table~\ref{table:no-seg-ablation}, removing segmentation consistently degrades performance: MME drops from \(1654/628\) to \(1620/585\) (perception/reasoning), MMBench falls by \(1.6\) points, and SEED-Bench declines by \(2.3\) points. These results indicate that while the model learns meaningful depth associations between adjacent objects and entities, depth alone is insufficient for strong VQA-style reasoning; explicit semantic grouping from segmentation complements geometry and yields superior accuracy.

\begin{wraptable}[4]{r}{0.52\columnwidth}
\vspace{-15pt}
\centering
\scriptsize 
\setlength{\tabcolsep}{3pt}
\caption{Ablation Study on Depth Tokens Effect on HardBLINK. Abbreviations: P = Perceptio; $-$Depth = w/o depth}
\begin{tabular}{l@{\hskip 6pt}cccc}
\toprule
Model & 3 Points & 4 Points & 5 Points & Avg. \\
\midrule
P & 75.8 & 71.0 & 66.1 & 71.0 \\
P ($-$Depth) & 48.4 & 48.4 & 38.7 & 45.2 \\
\bottomrule
\end{tabular}
\label{tab:hardBLINK}
\vspace{-5pt}
\end{wraptable}

\subsubsection{Impact of Loss Functions}
We further ablate the two depth-specific objectives introduced in Sec.~\ref{sec:method-loss-function}: the \emph{soft depth reconstruction} loss \(L_{\text{DepthRecon}}\) and the \emph{depth token generation} loss \(L_{\text{depth}}\). Removing \(L_{\text{DepthRecon}}\) reduces MME to \(1625/613\) and MMBench to 81.9\% (Table~\ref{tab:ablation-depth-loss-terms}); removing \(L_{\text{depth}}\) yields \(1632/621\), 82.4\% on MMBench, and a SEED-Bench drop from 75.7\% \(\to\) 74.3\%. Both objectives therefore contribute positively and in a complementary manner: \(L_{\text{DepthRecon}}\) strengthens continuous depth fidelity through differentiable decoding, while \(L_{\text{depth}}\) sharpens the discrete depth-token sequence itself. 



\begin{table}[t]
\centering

\begin{minipage}[t]{0.47\columnwidth}
\centering
\caption{Ablation Study on Perception Tokens Effect on benchmarks. Abbreviations: P = Perceptio; $-$Depth = w/o depth tokens; $-$Seg = w/o segmentation tokens.}
\label{tab:compact_results}
\resizebox{\linewidth}{!}{%
\begin{tabular}{lccc}
\hline
\textbf{Model} & \textbf{MME (perc./reas.)} & \textbf{MMBench (\%)} & \textbf{SEED-Bench (\%)} \\
\hline
P & 1654/628 & 83.4 & 75.7 \\
P ($-$Depth) & 1661/652 ($7/24\uparrow$)  & 83.8 ($0.4 \uparrow$) & 76.3 ($0.8 \uparrow$) \\
P ($-$Seg) & 1620/585 ($34/43\downarrow$)  & 81.8 ($1.6 \downarrow$) & 73.4 ($2.3 \downarrow$) \\
\hline
\end{tabular}%
}
\label{table:no-seg-ablation}
\end{minipage}
\hfill
\begin{minipage}[t]{0.49\columnwidth}
\centering
\caption{Ablation Study on Depth Loss terms on benchmarks. Abbreviations: P = Perceptio; $-$Depth-Rec = w/o depth reconstruction loss; $-$Depth-Gen = w/o depth token generation loss.}
\label{tab:ablation-depth-loss-terms}
\resizebox{\linewidth}{!}{%
\begin{tabular}{lccc}
\hline
\textbf{Model} & \textbf{MME (perc./reas.)} & \textbf{MMBench (\%)} & \textbf{SEED-Bench (\%)} \\
\hline
P & 1654/628 & 83.4 & 75.7 \\
P ($-$Depth-Rec) & 1625/613($29/15 \downarrow$) & 81.9 ($1.5 \downarrow$) & 73.7 ($2.0 \downarrow$)\\
P ($-$Depth-Gen) & 1632/621($22/7 \downarrow$) & 82.4 ($1.0 \downarrow$) & 74.3 ($1.6 \downarrow$) \\
\hline
\end{tabular}%
}
\end{minipage}

\end{table}



\section{Discussion and Conclusion}
Perceptio is a perception-enhanced LVLM that emits segmentation and discretized depth tokens inside the same autoregressive sequence, then generates text answers, turning 2D--3D perception into an in-sequence spatial chain-of-thought. This design, combined with composite depth-token and reconstruction objectives, materially strengthens spatial grounding in the trained model. Our trained model achieves state-of-the-art results on referring expression segmentation, depth understanding, and general VQA benchmarks.

Ours is, to our knowledge, the first work to jointly optimize 2D semantic segmentation and 3D depth reasoning within a single autoregressive LVLM sequence. We demonstrate strong performance on image-related perception and QA tasks. Prior work addresses these modalities in isolation; Perceptio closes this gap and demonstrates that the combination is necessary: removing depth tokens collapses HardBLINK accuracy by 25.8 points, while removing segmentation tokens degrades general VQA by up to 2.3 points on SEED-Bench.

\paragraph{Limitations.}
We identify limitations that suggest directions for future work.
First, our ablation (Table~\ref{tab:compact_results}) reveals a trade-off: removing depth tokens slightly \emph{improves} general VQA metrics (e.g., +0.4\% MMBench), suggesting that depth token generation introduces a mild optimization tension with text-only tasks. Mitigating this through task-adaptive curriculum learning is a promising direction.
Second, our training and evaluation are limited to static images; extending to video, where temporally consistent depth tokens and object tracking introduce new optimization challenges, remains open.
Third, Perceptio relies on frozen teacher models (Depth Anything V2, SAM2) whose errors propagate to the student; improving robustness to teacher noise is important for deployment.

\paragraph{Future Work.}
More broadly, we are motivated by the question of how models can learn generalizable spatial concepts that transfer across tasks and domains, for instance, extending perception tokens to encode surface normals or optical flow, moving toward a unified spatial intelligence within a single autoregressive framework.
%
%
\bibliographystyle{splncs04}
\bibliography{main}
\appendix

\section{Depth reconstruction loss}

The depth reconstruction loss is made differentiable by the soft token averaging method. The details of the algorithm flow are presented below in Algorithm \ref{alg:soft-mix}.

\begin{algorithm}
\caption{Soft codebook mixing for differentiable depth reconstruction}
\label{alg:soft-mix}
\begin{algorithmic}[1]
\Require Decoder logits \(\mathbf{z}\in\mathbb{R}^{B\times T\times V}\); ground-truth tokens \(\mathbf{y}\in\mathbb{N}^{B\times T}\); depth code index set \(\mathcal{D}\); codebook \(\{\mathbf{e}_k\}_{k\in\mathcal{D}}\).
\For{\(b=1\) \textbf{to} \(B\)}
  \State Parse depth span \((s_b,e_b)\) in \(\mathbf{y}_b\) with \(\mathbf{y}_{b,s_b}=d_{\text{start}}\), \(\mathbf{y}_{b,e_b}=d_{\text{end}}\)
  \If{\(s_b,e_b\) found}
    \State \(L_b \gets e_b - s_b - 1\)
    \For{\(t=s_b+1\) \textbf{to} \(e_b-1\)}
      \State \(\boldsymbol{\ell} \gets \mathbf{z}_{b,t-1}\) \Comment{logits predicting token at \(t\)}
      \State Mask non-depth indices: \(\boldsymbol{\ell}_{j}\!\leftarrow\!-\infty\) if \(j\notin\mathcal{D}\)
      \State \(p \gets \mathrm{softmax}(\boldsymbol{\ell})\) \Comment{\(p\in\mathbb{R}^{V}\) support on \(\mathcal{D}\)}
      \State \(\tilde{\mathbf{z}}_{b,t} \gets \sum_{k\in\mathcal{D}} p_k\,\mathbf{e}_k\)
    \EndFor
    \State truncate \(\{\tilde{\mathbf{z}}_{b,t}\}\) to \(n\), reshape to \(\sqrt{n}{\times}\sqrt{n}\)
    \State \(\hat{\mathbf{Y}}_b \gets \mathrm{VQVAE_{Decoder}}(\{\tilde{\mathbf{z}}_{b,t}\})\)
  \Else
    \State \(\hat{\mathbf{Y}}_b \gets \mathbf{0}\) \Comment{no valid depth span}
  \EndIf
\EndFor
\State Compute \(L_{\text{DepthRecon}} = \frac{1}{B}\sum_{b=1}^{B}\big\|\hat{\mathbf{Y}}_b - \mathbf{Y}_b\big\|_2^2\)
\end{algorithmic}
\end{algorithm}

\section{3D reasoning task examples}
\begin{figure}[t]
  \centering
    \begin{tabular}{c}
         \includegraphics[width=0.9\linewidth]{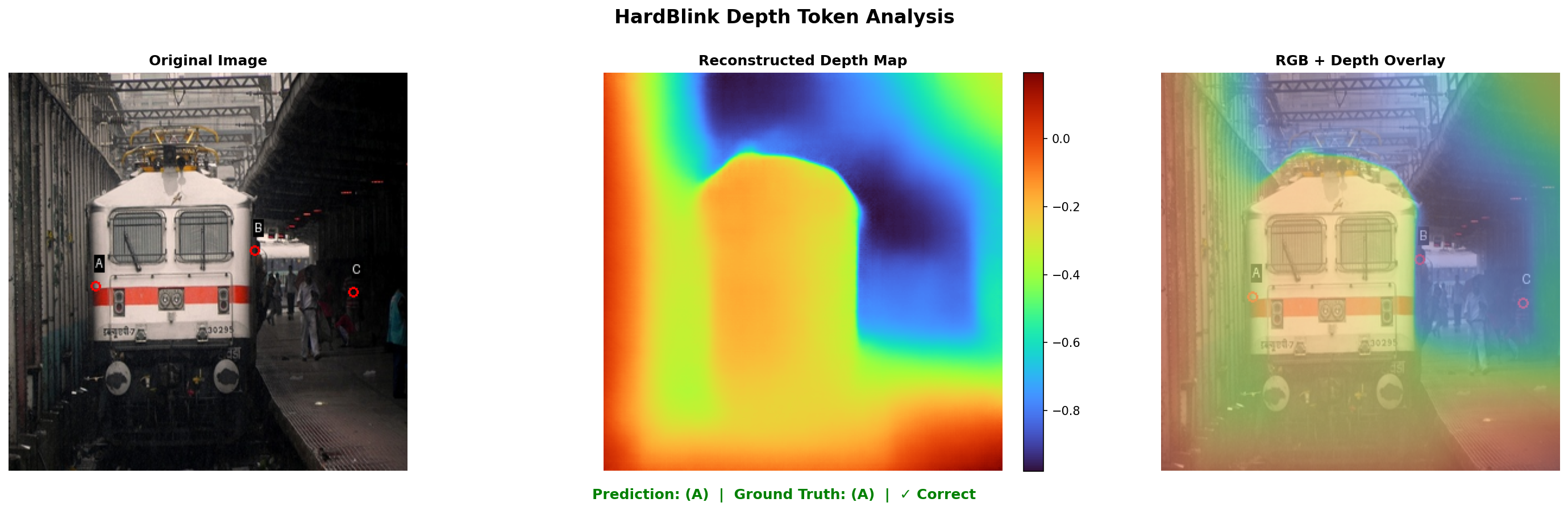} \\
         \includegraphics[width=0.9\linewidth]{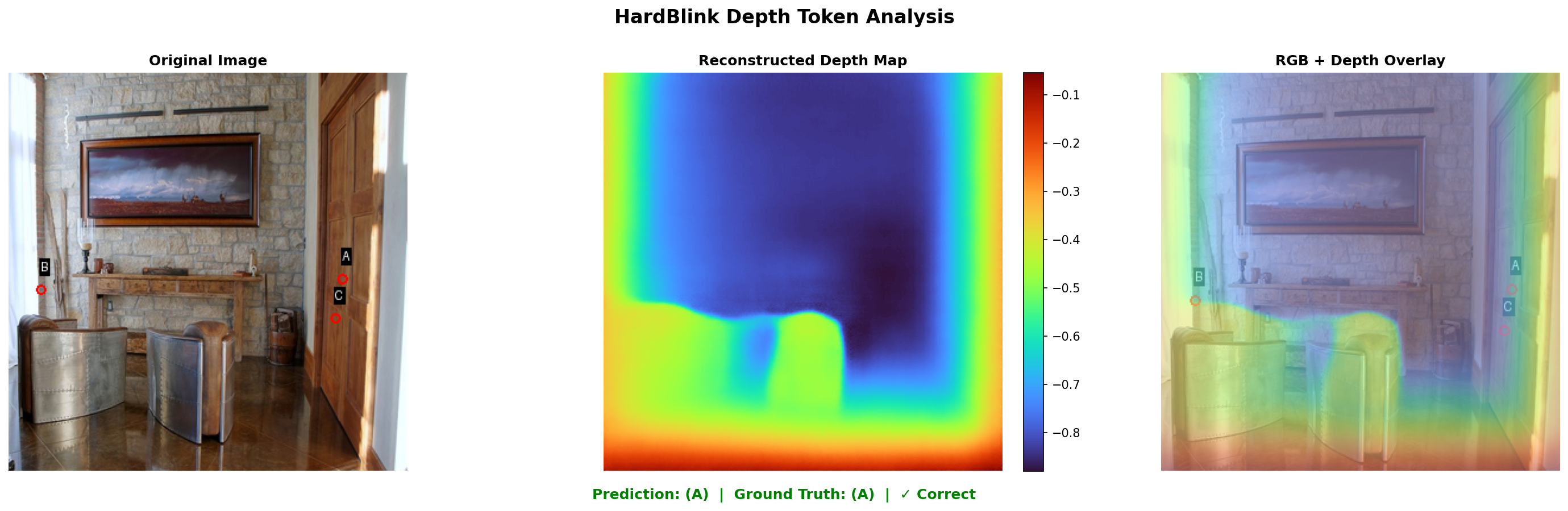} \\
         \includegraphics[width=0.9\linewidth]{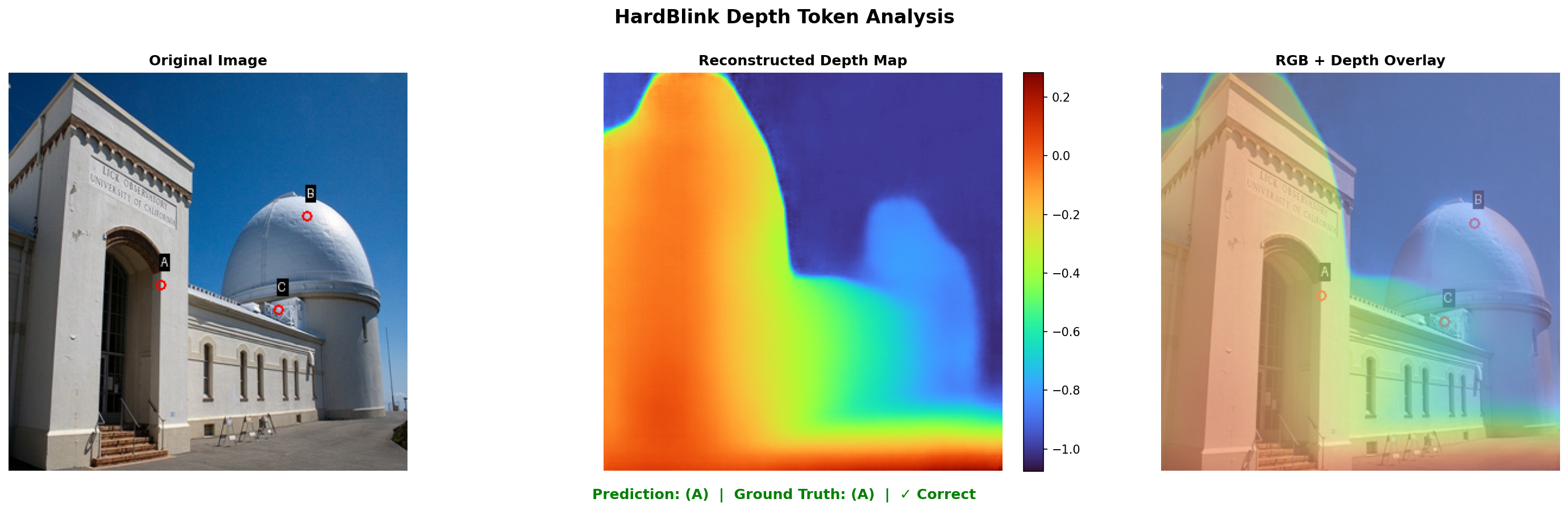} \\
         \includegraphics[width=0.9\linewidth]{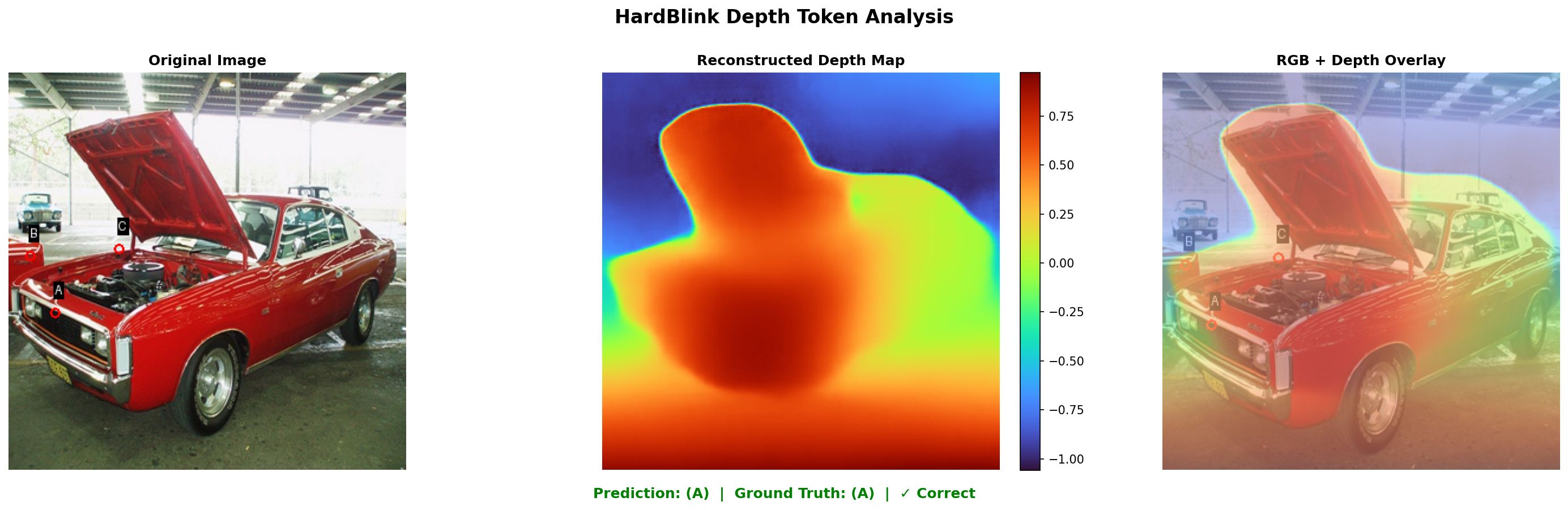} \\
    \end{tabular}
    \caption{$(left \rightarrow right)$ Hardblink dataset examples with original images (first column), reconstructed depth maps from our method (second column) and overlay with the original image (third column). The overlayed depth maps show that Perceptio model makes correct decision in alignment with the perceived depth map.}
    \label{fig:hardblink_samples}
\end{figure}

The reasoning task that requires 3D depth signal in the Hardblink dataset is presented in the Fig. \ref{fig:hardblink_samples}.

\section{2D reasoning task examples}
\begin{figure}[t]
  \centering
    \begin{tabular}{c}
         \includegraphics[width=0.7\linewidth]{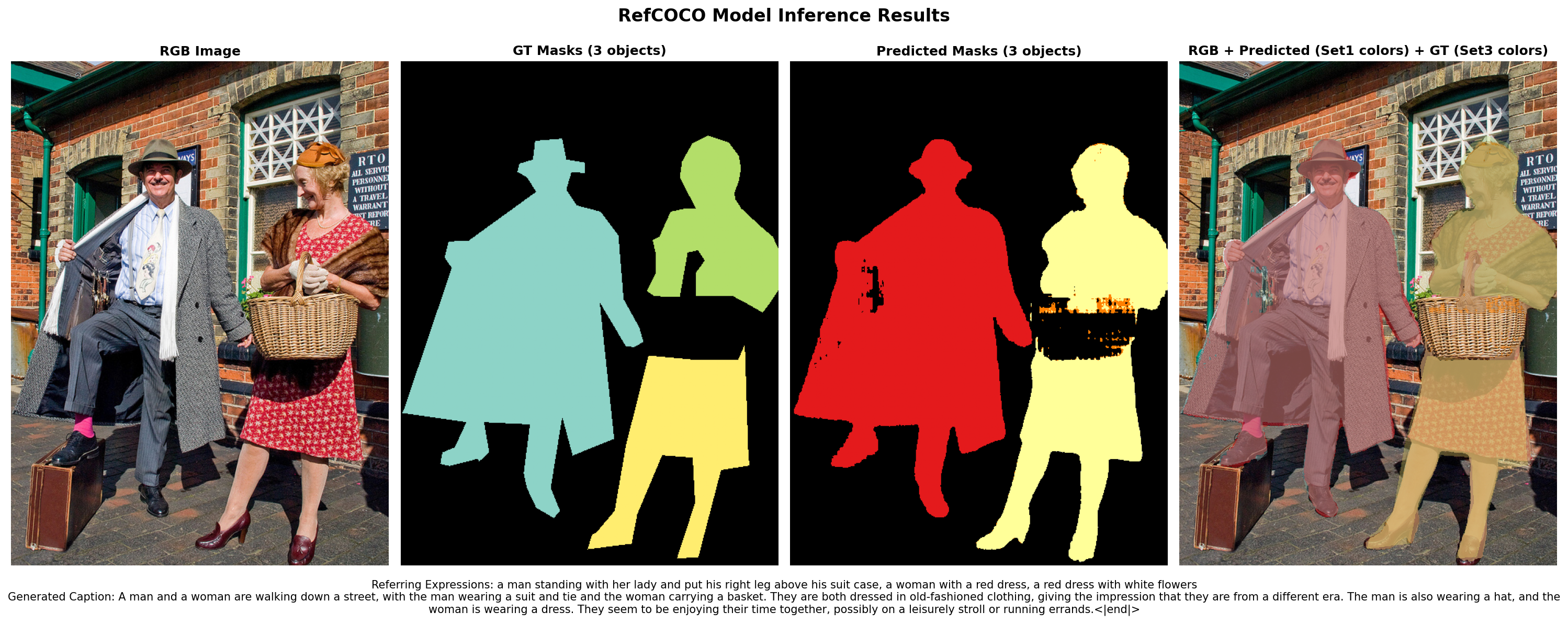} \\
         \includegraphics[width=0.7\linewidth]{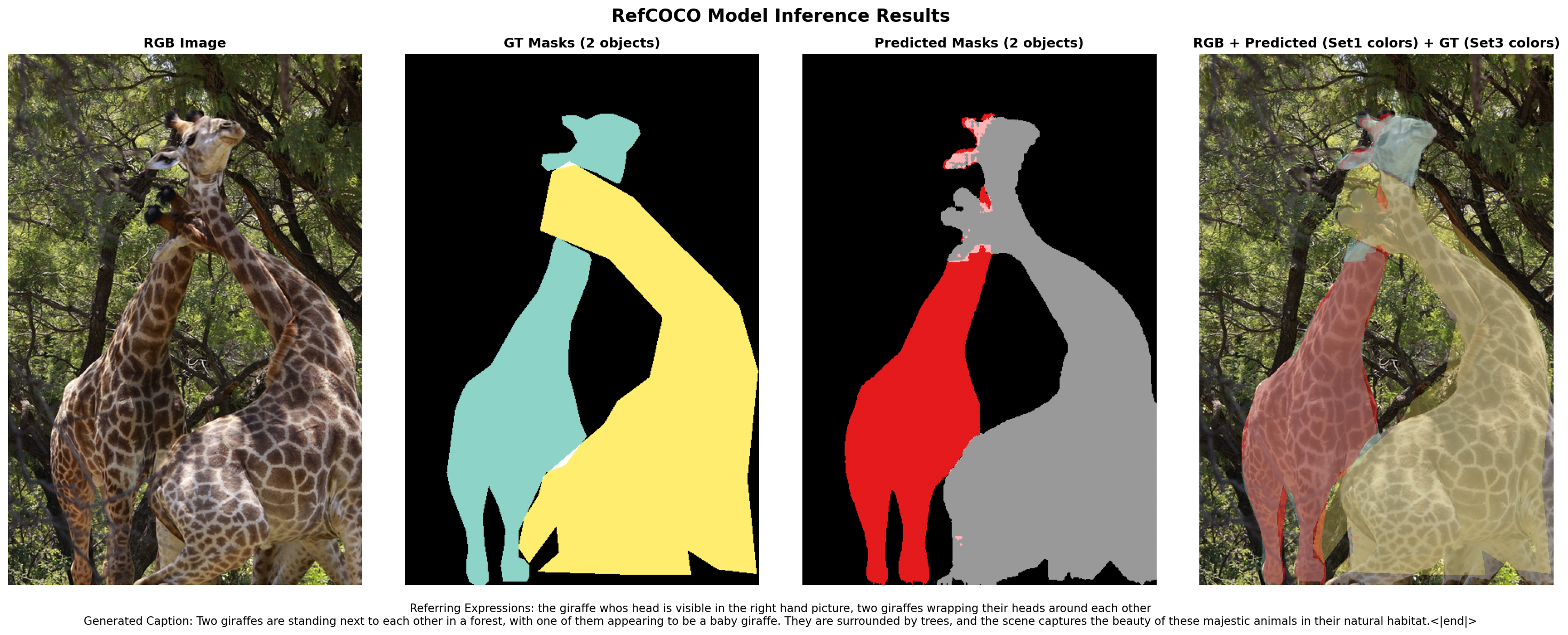} \\
         \includegraphics[width=0.7\linewidth]{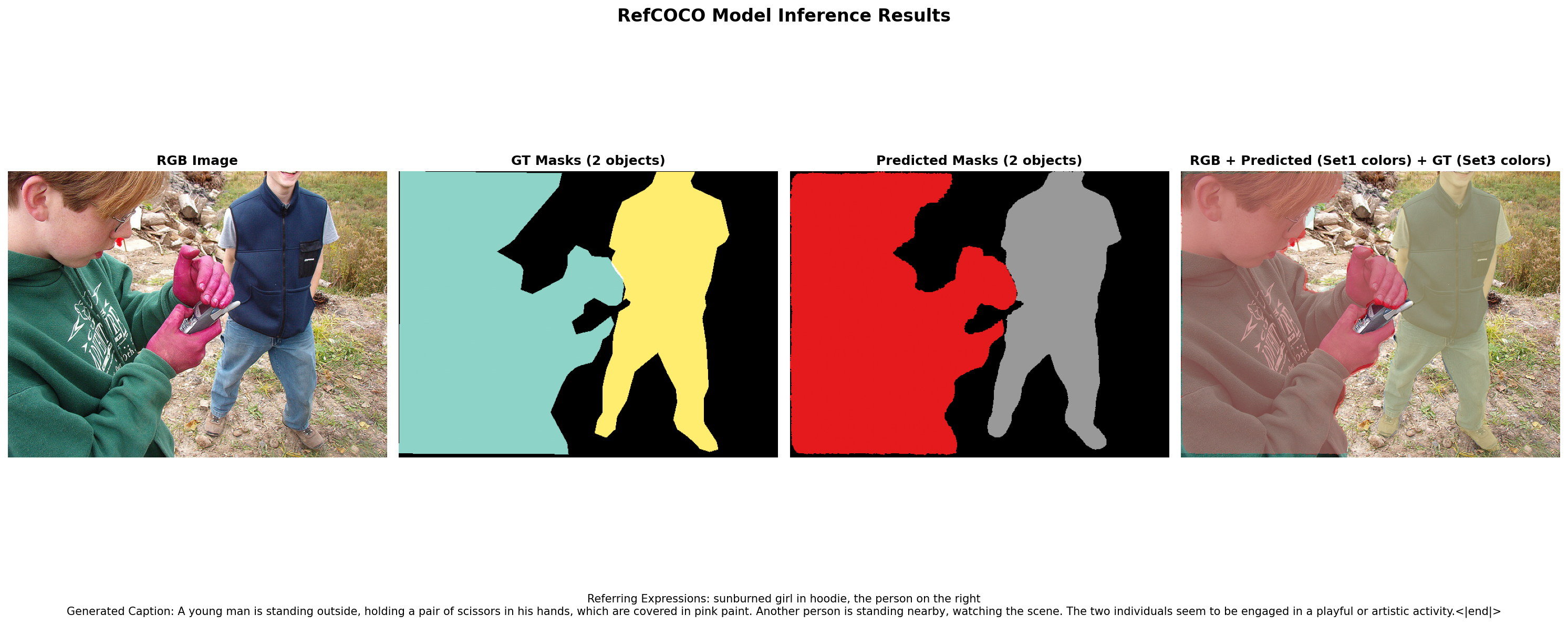} \\
         \includegraphics[width=0.7\linewidth]{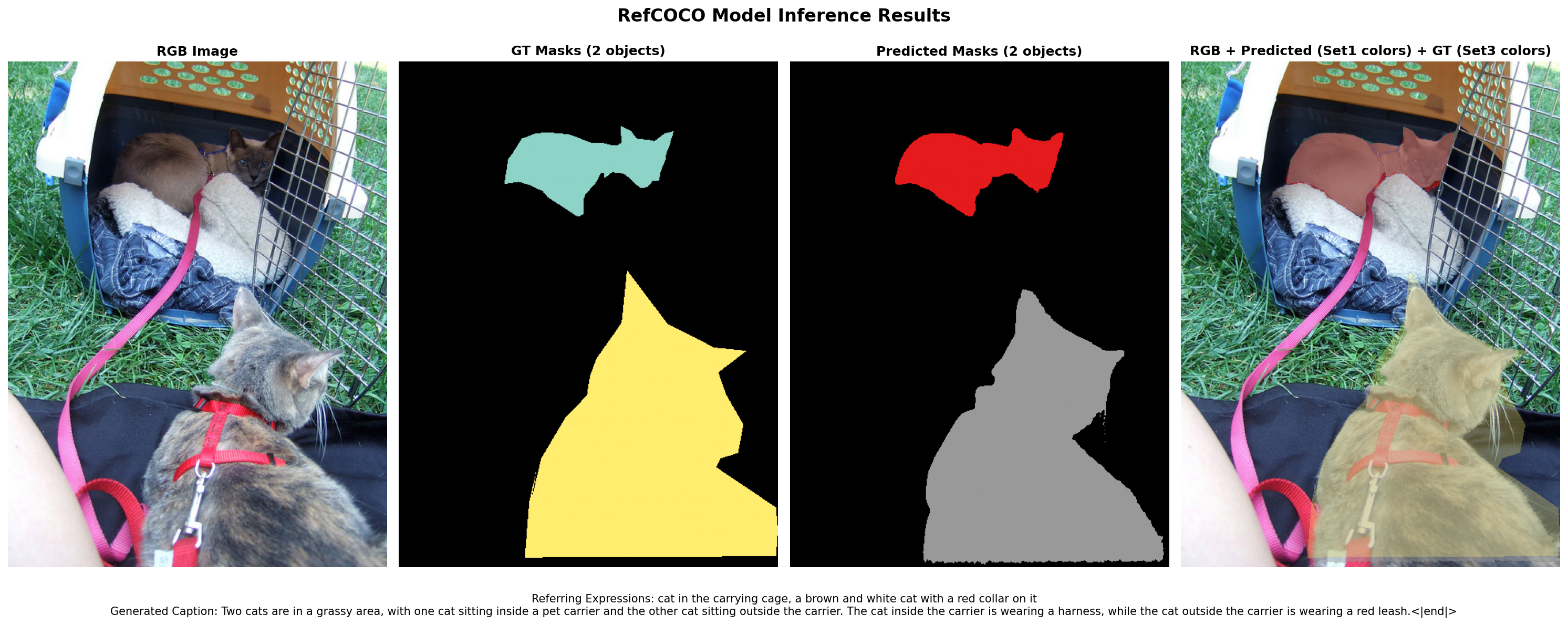} \\
    \end{tabular}
    \caption{$(left \rightarrow right)$ RefCoCoG dataset samples with the RGB image (first column), ground truth segmentation mask (second column), predicted segmentation mask (third column) and the overlayed predicted mask (fourth column) from the Perceptio model.}
    \label{fig:refcocog_samples}
\end{figure}

Examples of reasoning and segmentation mask predictions in the RefCOCOg dataset from Perceptio model predictions are shown in Fig. \ref{fig:refcocog_samples}.

\end{document}